\documentclass[conference]{IEEEtran}

\usepackage{cite}
\usepackage{amsmath,amssymb,amsfonts}
\usepackage{algorithmic}
\usepackage{graphicx}
\usepackage[caption=false,font=footnotesize]{subfig}
\usepackage{textcomp}
\usepackage{xcolor}
\usepackage{multirow}
\usepackage{verbatim}
\def\BibTeX{{\rm B\kern-.05em{\sc i\kern-.025em b}\kern-.08em
    T\kern-.1667em\lower.7ex\hbox{E}\kern-.125emX}}

\newif\ifcomments
\commentstrue     
\commentsfalse    

\ifcomments
    \newcommand{\shivaram}[1]{\textcolor{blue}{#1}}
    \newcommand{\saurbh}[1]{\textcolor{green!60!black}{#1}}
\else
    \newcommand{\shivaram}[1]{}
    \newcommand{\saurbh}[1]{}
\fi
    
\begin{document}

\title{Decentralized Multi-Robot Exploration with Probabilistic Peer Intent and Multi-hop Plan Propagation}

\author{
    \IEEEauthorblockN{Saurbh Singh Jamwal, Nived Chebrolu, Shivaram Kalyanakrishnan}
    \IEEEauthorblockA{\textit{Department of Computer Science and Engineering} \\
    \textit{Indian Institute of Technology Bombay}\\
    {saurbh, nived, shivaram}@cse.iitb.ac.in}
}


\maketitle

\begin{abstract}
Efficient coordination under limited communication remains a key challenge in decentralized multi-robot exploration. While centralized approaches benefit from global information sharing, they are often impractical in large-scale or communication-constrained environments. Existing Monte Carlo Tree Search (MCTS)-based approaches, such as Decentralized Monte Carlo Exploration (DMCE), enable decentralized planning by taking peer intent into account. This peer intent is obtained by communicating sequences of planned waypoints with robots within direct communication range. In this work, we extend this idea by introducing Probabilistic Peer Intent (PPI), which converts peer trajectories into a continuous spatial representation of predicted intent and incorporates it into local MCTS action evaluation. We additionally study the effects of sharing peer intent beyond direct communication range by propagating plans over multiple hops. Experiments across multiple simulated environments and team sizes show that PPI and Multi-hop propagation can each improve decentralized exploration, with their relative benefits depending on environment structure and team size. We also demonstrate the real-world deployment of our method on three robots operating in different environment types. \shivaram{Claim performance superiority in sim. as well as h/w.}
\end{abstract}

\begin{IEEEkeywords}
Multi-Robot Exploration, Decentralized Coordination, Monte Carlo Tree Search, Peer Intent, Multi-hop Communication
\end{IEEEkeywords}

\section{Introduction}

Autonomous exploration with multi-robot systems is a fundamental capability for time-critical applications such as search and rescue~\cite{queralta2020collaborative}, subterranean mapping~\cite{azpurua2023survey}, and planetary exploration~\cite{swinton2026evolution}. Decentralized coordination is particularly attractive in these settings due to its robustness to single points of failure and its ability to operate under limited and intermittent communication. A central challenge is to achieve efficient map coverage while limiting redundant sensing and traversal. At the same time, robots must maintain sufficient awareness of the team's exploration progress. \shivaram{Break long sentence.}\saurbh{done}

Recent approaches have moved beyond reactive frontier-based heuristics towards predictive planning methods that reason over future exploration actions~\cite{best2019decmcts,bone2023dmce}. Decentralized Monte Carlo Tree Search (MCTS)~\cite{best2019decmcts}  provides one such framework, allowing each robot to evaluate candidate trajectories through simulated sensing and evaluate information gain while incorporating communicated peer plans. Nevertheless, we identify two limitations when this framework is applied under restricted communication. First, representing peer intent only as sequences of planned waypoints does not explicitly capture how a peer's intended motion should influence nearby candidate paths. As a result, robots may independently select similar high-value regions despite having access to peer plans. Second, direct plan exchange is constrained by the instantaneous communication graph, limiting planning awareness as robots disperse~\cite{best2019decmcts,cladera2024opportunistic}. 

\begin{figure}[t]
\vspace{-6pt}
\centering
\includegraphics[width=\linewidth]{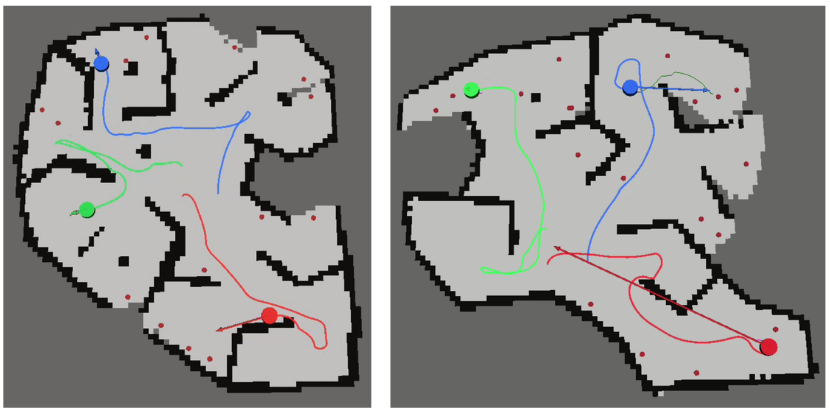}
\vspace{-10pt}
\caption{\textbf{Multi-robot exploration in two real-world indoor arenas.}
Online occupancy maps showing the regions explored during execution, together with the trajectories traced by the three robots.}
\label{fig:arena}
\vspace{-8pt}
\end{figure}

We address these limitations through two complementary extensions to decentralized MCTS. \textbf{(i) Probabilistic Peer Intent (PPI)} converts other robots trajectories into a continuous spatial representation of predicted peer intent and incorporates this information into local action evaluation.
\textbf{(ii) Multi-hop (MH) peer-plan propagation} relays trajectories from agents beyond the instantaneous one-hop neighborhood without requiring global connectivity or a central coordinator. This augments the planning information
available to each robot. 
PPI and Multi-hop propagation influence coordination through two distinct mechanisms. PPI modifies local action evaluation, whereas Multi-hop propagation changes the peer information available for distributed planning. 

We study their individual and combined effects across team sizes and environments ranging from open or branching spaces to cluttered and corridor-constrained tunnel environments. Our experiments show that both mechanisms can improve exploration individually, but their benefits are not uniformly additive. In constrained environments, increased peer-plan availability can also increase spatial intent exposure along shared passages,
negatively affecting the overall exploration goal.
This highlights an important distinction between extending the reach of coordination information and determining how that information should influence local planning.

The primary contributions of this paper are:
\begin{enumerate}
    \item We introduce Probabilistic Peer Intent (PPI), which converts communicated peer trajectories into a continuous spatial representation of predicted peer intent and incorporates it directly into the decentralized MCTS exploration objective.
    \item We introduce Multi-hop (MH) plan propagation, extending peer plan availability beyond directly communicating robots under range-limited and time-varying connectivity.
    \item We systematically evaluate PPI and MH across team sizes and environment structures, showing that both can improve exploration individually while their combination is topology dependent and not uniformly additive.
    \item We validate the proposed framework on three robots in previously unknown indoor environments, demonstrating the physical deployment of the same decentralized planning, PPI, and Multi-hop mechanisms used in simulation.
\end{enumerate}


The remainder of the paper is organized as follows. Section~II reviews related work, Section~III formulates the decentralized exploration problem, and Section~IV presents PPI and Multi-hop propagation. Section~V reports simulation and hardware results, followed by conclusion and future directions in Section~VI.
\shivaram{Need 2--3 lines giving organisation of paper. } \saurbh{done}

\section{Related Work}

Frontier-based exploration introduced by Yamauchi~\cite{yamauchi1997frontier} directs robots toward boundaries between known and unknown space and remains a widely used exploration strategy. Sampling-based methods, such as Multi-RRT exploration by Umari et al.~\cite{umari2017rrt}, instead sample trajectories through the environment rather than selecting only explicit frontier targets. In multi-robot settings, however, selecting exploration goals without coordination can lead to redundant coverage.

Several works extend exploration to multi-robot systems through explicit coordination. Yu et al.~\cite{yu2021smmr} propose SMMR-Explore, a submap-based framework combined with a multi-target potential field method to distribute robots across exploration goals. More broadly, multi-robot exploration systems commonly coordinate through target assignment, utility functions, or local interactions~\cite{wang2025survey}.  While these mechanisms can reduce redundant exploration, they primarily coordinate the selection or allocation of exploration targets rather than explicitly reasoning over the future trajectories of other robots.

Planning-based approaches explicitly reason over future robot actions. Best et al.~\cite{best2019decmcts} introduced Decentralized Monte Carlo Tree Search (Dec-MCTS), where agents coordinate by exchanging distributions over candidate plans, primarily for active perception tasks. Bone et al.~\cite{bone2023dmce} extended this idea to multi-robot exploration through Decentralized Monte Carlo Exploration (DMCE), where each robot performs local MCTS and replays communicated peer trajectories in simulation to account for their expected exploration. However, coordination remains limited to directly communicated trajectories, without explicitly representing spatial intent around the predicted paths. We address these limitations separately: PPI introduces a continuous spatial representation of available peer trajectories, while Multi-hop propagation extends peer-plan availability beyond direct neighbors.

Communication-constrained coordination has also been studied independently of exploration planning. MOCHA~\cite{cladera2024opportunistic} uses gossip-based Multi-hop communication to relay information beyond direct contacts, but operates primarily at the communication layer rather than integrating propagated information into exploration planning. Bramblett and Bezzo~\cite{bramblett2023epistemic} address communication loss through epistemic belief updates and task allocation, while other approaches reduce communication requirements through low-bandwidth position sharing~\cite{bayer2026lowbandwidth} or wireless signal sensing~\cite{jadhav2024wiserx}. In contrast, our Multi-hop mechanism propagates planned trajectories and makes non-local peer plans directly available to the decentralized exploration planner.

Recent work has also explored learning-based approaches for multi-robot coordination. MARVEL~\cite{chiun2025marvel} uses multi-agent reinforcement learning to learn decentralized exploration policies, while Meng et al.~\cite{meng2025gnnvae} employ graph neural networks to generate coordinated multi-agent behaviors. These approaches learn coordination policies from data, whereas our work focuses on model-based online planning. Specifically, we augment decentralized MCTS with an explicit spatial representation of communicated peer intent, allowing its influence on planning decisions to be directly controlled and analyzed without policy training.

Overall, existing approaches address complementary aspects of multi-robot coordination, including exploration-goal allocation, future-plan reasoning, and communication-constrained operation. Building on decentralized MCTS, we study two distinct extensions: PPI provides a continuous spatial representation of communicated peer trajectories for local planning, while Multi-hop propagation extends the availability of peer plans beyond direct neighbors. This separation allows us to examine how the representation and reach of coordination information independently affect multi-robot exploration. \shivaram{Contents okay, but make text less verbose, more succinct. Section can be compressed to 75\% of current.} \saurbh{Compressed it.}

\section{Problem Formulation}

\subsection{Decentralized Multi-Robot Exploration as an MDP}
\label{sec:problem_mdp}

We consider a team of $N$ robots exploring an initially unknown planar environment represented by an occupancy-grid map. Each robot maintains a local map from its sensor observations and makes exploration decisions using locally available information. For robot $i$, we formulate the finite-horizon exploration problem as an MDP

\begin{equation}
    \mathcal{M}^i = \left(\mathcal{S}^i,\mathcal{A}^i,T^i,R^i\right).
\end{equation}

The state $s_k^i \in \mathcal{S}^i$ consists of the robot position $p_k^i=(x_k^i,y_k^i)$ and its current local occupancy map $M_k^i$. At the exploration-planning level, robot orientation and motion dynamics are abstracted away, and actions specify feasible spatial waypoints. The displacement action space is continuous in direction, with each action specifying a fixed-length motion from the current position, while frontier locations provide additional goal-directed actions. An action is feasible if the resulting waypoint lies within the map and can be reached through known free space.

The transition function $T^i$ describes the evolution from the current state to the state associated with the selected waypoint, together with the corresponding update of the robot's local map from new observations. A sequence of actions over a finite horizon $H$ results in a trajectory

\begin{equation}
    \tau^i = \left(p_0^i,p_1^i,\ldots,p_H^i\right).
\end{equation}

The reward $R^i$ quantifies the exploration utility of a trajectory. The objective for each robot is to select a feasible trajectory that maximizes this reward,

\begin{equation}
    \tau^{i*} = \arg\max_{\tau^i} R^i(\tau^i).
\end{equation}

Since decision making is decentralized, each robot operates using its local map and information received from other robots. Under range-limited and intermittent communication, this information may be incomplete, which can lead to different robots exploring overlapping regions.

\shivaram{This subsection mixes the problem with the solution. Problem---the MDP formulation---needs to be made more precise. What is a trajectory?--sequence of points? There are infinitely many such; how did they get selected as actions?} \saurbh{Updated based on discussion}

\subsection{Assumptions}

We assume that all robots are localized in a common reference frame and that the localization error remains sufficiently small for exchanged positions and trajectories to be interpreted consistently, as in prior decentralized exploration systems~\cite{bartolomei2023fast,bayer2026lowbandwidth}. All explorable regions are assumed to be reachable, while low-level collision avoidance is handled independently of the high-level exploration planner. Communication is opportunistic and range-limited, resulting in intermittent local connectivity without requiring a globally connected communication network~\cite{best2019decmcts,cladera2024opportunistic}. \shivaram{Cite some application studies to justify why these assumptions are reasonable.}\saurbh{done using assumptions made in similar papers}

\section{Methodology}

\subsection{Baseline: Decentralized MCTS-Based Exploration}

\shivaram{Move this subsection to methodology. Methodology section needs more illustration, technical rigour. Ideally, you should show pseudocode for the baseline with 1--2 modules that are default in the baseline, but refined upon in this paper (PPI and multihop).}\saurbh{Moved. The two components are discussed in next 2 subsections.}

We build upon Decentralized Monte Carlo Exploration (DMCE)~\cite{bone2023dmce}, where each robot independently performs MCTS to select exploration actions using its locally available map and peer information. Coordination is introduced by exchanging planned trajectories between directly communicating robots. During planning, received peer trajectories are sampled and replayed through simulated sensing, updating the predicted map to account for regions that peers are expected to explore before evaluating the local robot's actions.

We retain the underlying DMCE planning procedure and introduce two extensions to peer plan coordination, detailed in the following subsections: \textbf{(i) Probabilistic Peer Intent (PPI)}, which modifies how available peer plans influence local action evaluation; and \textbf{(ii) Multi-hop propagation (MH)}, which extends peer plan availability beyond direct communication neighbors. In this work, we evaluate peer-intent representation and information reach separately and in combination.

\subsection{Probabilistic Peer Intent (PPI)}
\label{sec:ppi}

\begin{figure*}[t]
    \centering
    \includegraphics[width=\textwidth]{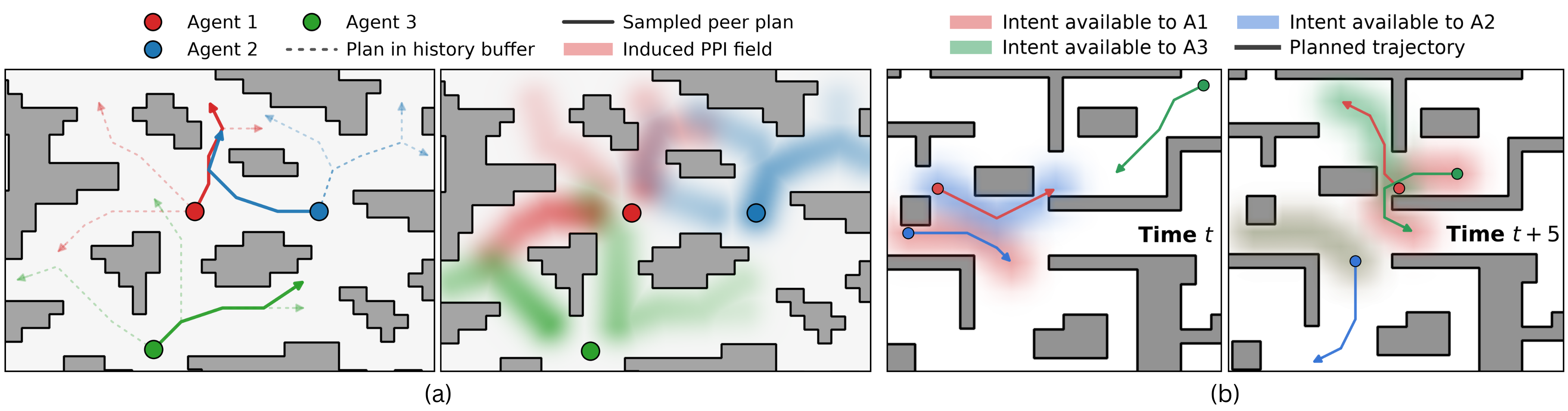}
    \caption{\textbf{Overview of the proposed coordination mechanisms.}
    \textbf{(a)} Probabilistic Peer Intent (PPI), where communicated peer trajectories are transformed into spatial intent fields used during local planning.
    \textbf{(b)} Multi-hop (MH) plan propagation, which extends peer-plan information beyond direct communication neighbors by relaying fresh peer plans while preserving their source identity and generation time.}
    \label{fig:method_overview}
\end{figure*}

DMCE incorporates a communicated peer trajectory by replaying the peer's predicted motion and sensing into the simulated map, thereby accounting for regions that the peer is expected to explore. However, this mechanism does not explicitly quantify the spatial interaction between a local candidate trajectory and the peer's predicted path. We therefore introduce Probabilistic Peer Intent (PPI), which transforms each available peer trajectory into a continuous spatial field. As illustrated in Fig.~\ref{fig:method_overview}(a), trajectories passing close to predicted peer motion receive greater exposure, while spatially separated trajectories receive weaker influence.

For each peer $j$, we retain its most recent valid trajectory and discard plans older than a freshness threshold $T_c$. Consecutive waypoints are linearly interpolated with spacing $\Delta_p$ to provide spatial support independent of the original waypoint spacing. Let $\mathbf{q}_{k,l}^{j}\in\mathbb{R}^2$ denote the $l$-th interpolated support point between future waypoints $k$ and $k+1$ of peer $j$. Its influence at location $\mathbf{x}$ is
\begin{equation}
P_{k,l}^{j}(\mathbf{x})=
\exp(-\lambda_s k)
\exp\left(-\frac{\|\mathbf{x}-\mathbf{q}_{k,l}^{j}\|^2}{2\sigma^2}\right),
\label{eq:ppi_kernel}
\end{equation}
where $\sigma$ controls the spatial extent of peer intent and $\lambda_s$ discounts positions farther along the communicated trajectory. The PPI field available to robot $i$ is then
\begin{equation}
P_i(\mathbf{x})=\max_{j,k,l}P_{k,l}^{j}(\mathbf{x}), \qquad P_i(\mathbf{x})\in[0,1].
\label{eq:ppi_field}
\end{equation}

The Gaussian kernel provides smoothly decreasing influence around predicted paths, while future-step decay gives greater importance to near-term peer motion. Max aggregation retains the strongest predicted intent at each location without increasing the field magnitude due to dense interpolation or overlapping peer trajectories. We use $\lambda_s=0.02$, $T_c=5.0$\,s, and $\Delta_p=0.10$\,m throughout the experiments.

\subsection{PPI-based Reward Integration}

The underlying DMCE planner evaluates each MCTS rollout according to the information expected to be acquired through simulated sensing. Let $M_0^i$ and $M_k^i$ denote the simulated occupancy maps at the beginning of a rollout and after step $k$, respectively. The explored-map value is
\begin{equation}
V(M_k^i)=1-H_{\mathrm{rel}}(M_k^i),
\end{equation}
where $H_{\mathrm{rel}}(\cdot)$ is the relative map entropy. The resulting information gain is
\begin{equation}
\mathrm{IG}(s_k^i)=V(M_k^i)-V(M_0^i).
\label{eq:information_gain}
\end{equation}

PPI augments this rollout evaluation by penalizing candidate trajectories that traverse regions of high predicted peer intent. For a candidate trajectory $\tau^i$, cumulative peer-intent exposure is defined as
\begin{equation}
C_{\mathrm{PPI}}(\tau^i)=\sum_{k=1}^{H}P_i(\mathbf{p}_k^i),
\label{eq:ppi_exposure}
\end{equation}
where $\mathbf{p}_k^i$ is the robot position at rollout step $k$. The resulting rollout objective becomes
\begin{equation}
R(\tau^i)=\mathrm{IG}(s_H^i)-\lambda_{\mathrm{PPI}}C_{\mathrm{PPI}}(\tau^i),
\label{eq:ppi_reward}
\end{equation}
where $\lambda_{\mathrm{PPI}}$ balances expected information gain against exposure to predicted peer intent.

Only the local robot's candidate trajectory is evaluated using PPI. Communicated peer trajectories continue to serve their original role in DMCE by predicting future map observations during rollout simulation, while the PPI penalty evaluates how the local trajectory relates to the resulting peer-intent field. Thus, PPI complements the original peer-plan reasoning by modifying the rollout objective rather than the underlying map prediction process.

\subsection{Multi-hop Information Propagation}

The original DMCE framework exchanges planned trajectories only between robots within direct communication range, limiting peer-plan availability to one-hop neighbors. Our Multi-hop (MH) extension allows each robot to forward both its own trajectory and previously received peer plans, enabling information to propagate across successive communication encounters as

\begin{equation}
\pi_j \rightarrow r_1 \rightarrow \cdots \rightarrow r_m \rightarrow r_i,
\end{equation}
where $\pi_j$ is generated by robot $r_j$, and each relay preserves the original source identity and generation timestamp.

For each source robot, only the most recent received plan is retained. Because relayed plans keep their original timestamps, stale information is naturally discarded using the same timeout  $T_c$ defined in Section~\ref{sec:ppi}. Figure~\ref{fig:method_overview}(b) illustrates this process: a robot can carry a previously received peer plan into a new communication neighborhood and relay it to a robot that never directly communicated with the original source. MH therefore extends peer-plan availability without changing the underlying MCTS planning or reward formulation.

\section{Results and Discussion}

\begin{table*}[t]
\centering
\caption{Comparison with other methods across environments using 5 robots, averaged over 10 independent runs. Lower is better. Runs are limited to $450$\,s; ``--'' indicates that the corresponding coverage threshold was not reached in all 10 runs.}
\label{tab:main_baselines}
\footnotesize
\setlength{\tabcolsep}{3.5pt}
\renewcommand{\arraystretch}{1.0}
\begin{tabular}{l | ccc | ccc | ccc | ccc}
\hline
& \multicolumn{3}{c|}{\textbf{Open}}
& \multicolumn{3}{c|}{\textbf{Forest}}
& \multicolumn{3}{c|}{\textbf{Urban}}
& \multicolumn{3}{c}{\textbf{Tunnel}} \\
\textbf{Method}
& $\textbf{T90}\downarrow$ & $\textbf{T95}\downarrow$ & $\textbf{Overlap}\downarrow$
& $\textbf{T90}\downarrow$ & $\textbf{T95}\downarrow$ & $\textbf{Overlap}\downarrow$
& $\textbf{T90}\downarrow$ & $\textbf{T95}\downarrow$ & $\textbf{Overlap}\downarrow$
& $\textbf{T90}\downarrow$ & $\textbf{T95}\downarrow$ & $\textbf{Overlap}\downarrow$ \\
\hline

MCTS (Uncoord.)
& 150.9 & 177.0 & 49.46
& 94.9 & 111.4 & 60.41
& 162.0 & 178.6 & 58.64
& 97.7 & 107.5 & 56.03 \\

RRT
& 183.1 & 231.2 & 20.88
& 87.3 & 130.1 & 40.50
& -- & -- & 44.48
& 317.9 & 353.5 & 35.38 \\

MMPF
& -- & -- & 48.14
& 107.2 & 123.4 & 54.32
& -- & -- & 69.58
& -- & -- & 70.11 \\

DMCE
& 132.1 & 150.9 & 32.59
& 92.0 & 117.0 & 49.95
& 158.9 & 169.4 & 53.99
& 85.5 & 100.4 & 50.77 \\

DMCE-Global
& \textbf{89.7} & \textbf{107.8} & 35.65
& \textbf{69.4} & \textbf{81.3} & 51.85
& \textbf{132.3} & \textbf{157.7} & 49.45
& \textbf{70.5} & \textbf{79.6} & 56.75 \\

MARVEL
& 228.5 & 288.8 & \textbf{9.42}
& 134.3 & 154.6 & \textbf{20.37}
& -- & -- & \textbf{33.23}
& -- & -- & \textbf{20.57} \\

\hline

\textbf{DMCE +PPI}
& 132.9 & 161.3 & 26.22
& 73.3 & 94.5 & 47.64
& 170.3 & 179.5 & 53.42
& 95.7 & 106.4 & 48.92 \\

\textbf{DMCE +MH}
& 121.1 & 154.3 & 38.74
& 71.2 & 93.2 & 48.58
& 154.1 & 170.7 & 58.51
& 95.7 & 103.2 & 49.74 \\

\textbf{DMCE +PPI +MH}
& 139.1 & 167.4 & 22.62
& 79.0 & 95.1 & 44.08
& 159.0 & 177.0 & 51.26
& 109.7 & 117.4 & 49.72 \\

\hline
\end{tabular}
\end{table*}

\begin{table*}[t]
\centering
\caption{Component analysis of PPI and Multi-hop propagation across team sizes. Lower is better.}
\label{tab:ablation}
\footnotesize
\setlength{\tabcolsep}{8.5pt}
\renewcommand{\arraystretch}{1.0}
\begin{tabular}{c l | cc | cc | cc}
\hline
& &
\multicolumn{2}{c|}{\textbf{Forest}} &
\multicolumn{2}{c|}{\textbf{Urban}} &
\multicolumn{2}{c}{\textbf{Tunnel}} \\
\textbf{Robots} & \textbf{Method}
& $\textbf{T95}\downarrow$ & $\textbf{Overlap}\downarrow$
& $\textbf{T95}\downarrow$ & $\textbf{Overlap}\downarrow$
& $\textbf{T95}\downarrow$ & $\textbf{Overlap}\downarrow$ \\
\hline

\multirow{4}{*}{5}
& DMCE
& $117.0{\pm}30.6$ & $49.95{\pm}6.75$
& $\mathbf{169.4{\pm}19.9}$ & $53.99{\pm}6.53$
& $\mathbf{100.4{\pm}16.1}$ & $50.77{\pm}4.96$ \\

& \textbf{+PPI}
& $94.5{\pm}18.8$ & $47.64{\pm}8.30$
& $179.5{\pm}16.8$ & $53.42{\pm}8.41$
& $106.4{\pm}11.3$ & $\mathbf{48.92{\pm}7.23}$ \\

& \textbf{+MH}
& $\mathbf{93.2{\pm}17.6}$ & $48.58{\pm}7.47$
& $170.7{\pm}13.7$ & $58.51{\pm}5.76$
& $103.2{\pm}13.4$ & $49.74{\pm}8.87$ \\

& \textbf{+PPI+MH}
& $95.1{\pm}11.7$ & $\mathbf{44.08{\pm}8.06}$
& $177.0{\pm}16.9$ & $\mathbf{51.26{\pm}10.80}$
& $117.4{\pm}10.5$ & $49.72{\pm}6.59$ \\

\hline

\multirow{4}{*}{8}
& DMCE
& $\mathbf{63.8{\pm}9.2}$ & $62.39{\pm}7.08$
& $154.9{\pm}20.8$ & $\mathbf{70.26{\pm}5.65}$
& $\mathbf{91.7{\pm}5.2}$ & $\mathbf{60.46{\pm}4.86}$ \\

& \textbf{+PPI}
& $64.9{\pm}7.1$ & $\mathbf{55.91{\pm}4.33}$
& $180.0{\pm}12.3$ & $73.67{\pm}4.08$
& $97.3{\pm}8.7$ & $64.05{\pm}2.86$ \\

& \textbf{+MH}
& $67.7{\pm}8.1$ & $64.93{\pm}3.44$
& $\mathbf{153.0{\pm}18.2}$ & $73.35{\pm}2.82$
& $94.6{\pm}7.7$ & $65.97{\pm}5.34$ \\

& \textbf{+PPI+MH}
& $\mathbf{63.9{\pm}7.1}$ & $57.11{\pm}6.26$
& $177.8{\pm}24.5$ & $73.42{\pm}3.19$
& $104.8{\pm}7.6$ & $62.48{\pm}4.53$ \\

\hline

\multirow{4}{*}{12}
& DMCE
& $59.5{\pm}4.2$ & $74.10{\pm}3.13$
& $171.6{\pm}15.3$ & $\mathbf{79.73{\pm}3.53}$
& $106.4{\pm}9.5$ & $74.52{\pm}4.09$ \\

& \textbf{+PPI}
& $\mathbf{55.0{\pm}3.9}$ & $\mathbf{69.10{\pm}2.68}$
& $168.3{\pm}16.4$ & $83.15{\pm}3.62$
& $99.4{\pm}9.3$ & $\mathbf{73.19{\pm}3.32}$ \\

& \textbf{+MH}
& $55.4{\pm}4.6$ & $73.16{\pm}2.14$
& $\mathbf{166.0{\pm}18.6}$ & $83.69{\pm}2.71$
& $\mathbf{89.6{\pm}8.3}$ & $74.31{\pm}3.38$ \\

& \textbf{+PPI+MH}
& $62.0{\pm}4.6$ & $70.42{\pm}3.67$
& $183.8{\pm}16.7$ & $82.90{\pm}1.45$
& $108.7{\pm}12.2$ & $75.42{\pm}3.33$ \\

\hline
\end{tabular}
\end{table*}

\subsection{Experimental Setup}

\begin{figure*}[!htbp]
\centering
\includegraphics[width=0.95\textwidth]{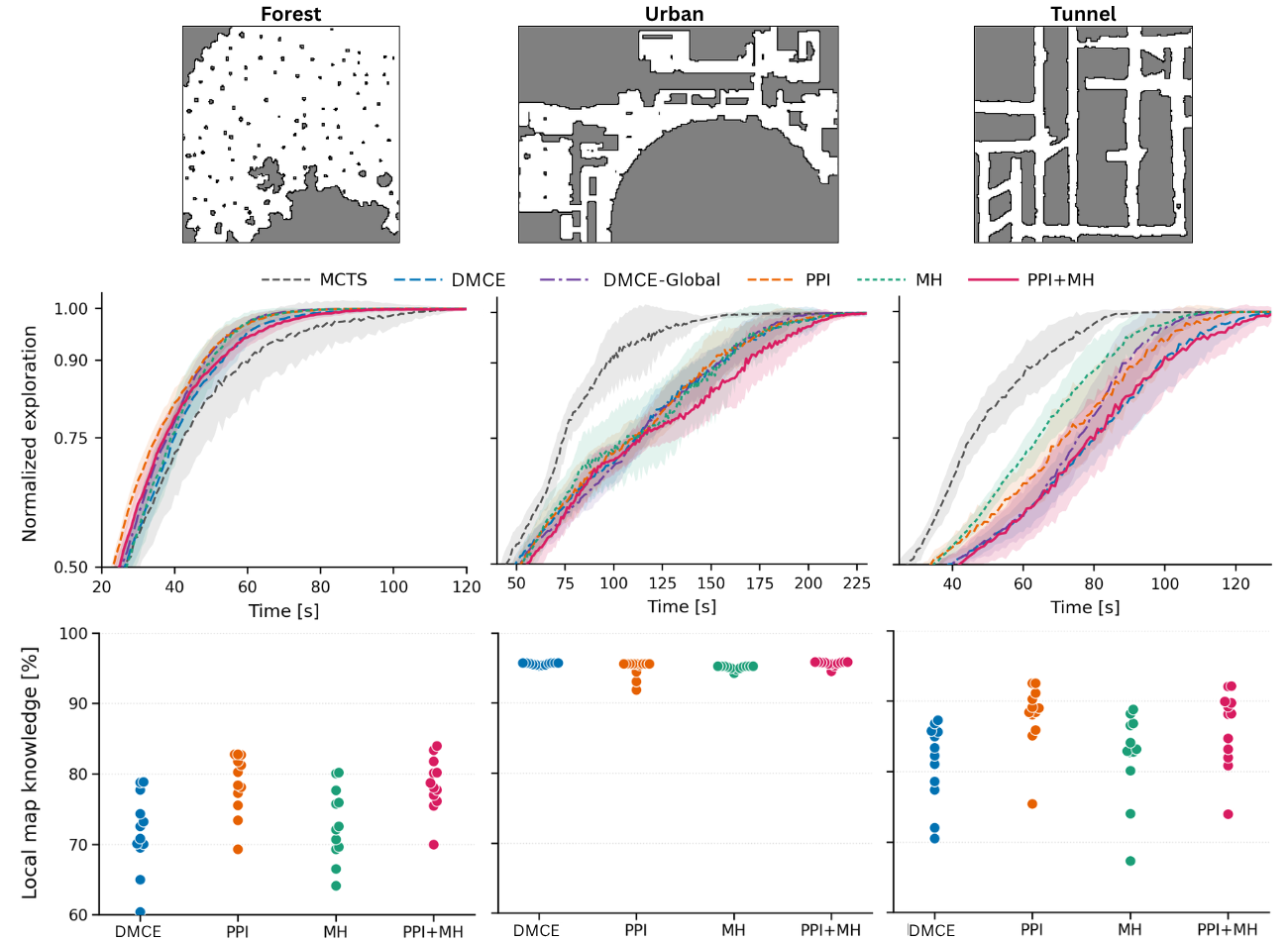}
\vspace{-2pt}
\caption{Exploration behavior and map knowledge distribution. Experiments performed with 12 robots in (a) Forest, (b) Urban, and (c) Tunnel. \textbf{Top:} environment layouts. \textbf{Middle:} normalized coverage over time, with shaded regions indicating variation across 10 independent runs. \textbf{Bottom:} local map knowledge at T95 for the DMCE variants. Each particle represents the average knowledge of one within-run agent computed across the 10 runs. Agents are ranked from least- to best-informed within each run before averaging. 
}
\label{fig:multi_map}
\vspace{-2pt}
\end{figure*}

We evaluate our approach in simulation across four environments: \textit{Open}, \textit{Forest}, \textit{Urban}, and \textit{Tunnel}. \textit{Open} is evaluated with a team size of $N=5$, while the remaining environments use $N=\{5,8,12\}$. As visualized in Fig.~\ref{fig:multi_map}, these environments range from open spaces to more branched and corridor-like layouts.
We evaluate both the effects of PPI and Multi-hop propagation (MH) strategies individually as well as jointly.  $\lambda_{\mathrm{PPI}}=1$ and $\sigma=1$, with an $8$\,m sensor range for mapping and $15$\,m line-of-sight communication range are used throughout the experiments. Our primary evaluation metrics are coverage time (T90 and T95) and lifetime overlap, i.e., the fraction of the total explored area that is redundantly mapped by multiple robots (i.e., at least by two robots). Methods that have a lower time-to-coverage and lifetime overlap are considered better.
Results are reported as the mean and standard deviation across 10 
independent runs across methods. 
\shivaram{Refer back to Figure 1 for the visualisation.}

\subsection{Performance Analysis}

In this section, we compare our proposed extensions (PPI and MH map propagation) against exploration methods including uncoordinated, sampling-based, decentralized, globally connected, and learned approaches. \textbf{MCTS (Uncoord.)} performs independent Monte Carlo tree search without peer-plan coordination, \textbf{RRT}~\cite{umari2017rrt} provides a sampling-based baseline, and \textbf{MMPF}~\cite{yu2021smmr} uses multi-robot potential fields. \textbf{DMCE}~\cite{bone2023dmce} is our primary decentralized baseline and operates under the same $15$\,m line-of-sight communication constraint as our extensions, while \textbf{DMCE-Global} provides a reference with globally available peer plans. We compare against \textbf{MARVEL}~\cite{chiun2025marvel} as a learned multi-robot exploration baseline.

Table~\ref{tab:main_baselines} compares the methods using $N=5$ robots. Relative to DMCE, our proposed extensions exhibit environment-dependent trade-offs between coverage time and redundant exploration. PPI substantially reduces overlap in open environments and improves coverage time in forest environments. We observe that the MH strategy provides a similar improvement in the forest environment. By combining both PPI and MH, we observe the lowest overlap among the DMCE variants in open, forest, and urban environments, however, the coverage time does not improve consistently. We note that DMCE-Global (which has access to all robot plans at all times) consistently performs well across all environments, illustrating the benefit of unrestricted peer-plan availability. In the next sections, we examine the individual contributions, scaling behavior, and interaction of PPI and MH.

\subsection{Analysis of PPI and Multi-hop Strategies}
\label{sec:component_analysis}

Table~\ref{tab:ablation} isolates the effects of PPI and Multi-hop propagation across team sizes. In Forest, both strategies consistently improve exploration time. For $N=5$, PPI and MH reduce T95 from $117.0$\,s to $94.5$\,s and $93.2$\,s, respectively, while PPI+MH achieves the lowest overlap. Similar individual gains persist with $N=12$, which indicates the benefits of both spatial intent reasoning and extended peer-plan availability as the team disperses in different directions.

The effects are more topology-dependent in Urban and Tunnel. At $N=12$, uncoordinated MCTS is highly competitive in coverage time, potentially because of higher robot density relative to the available branches and passages. This can induce spatial separation even without explicit coordination. In such settings, environment geometry itself can partially distribute robots across distinct routes, reducing the immediate advantage of explicit peer-plan reasoning. Among coordinated methods, both PPI and MH improve T95 over DMCE, with MH providing the largest reduction in Tunnel ($106.4$ to $89.6$\,s), also outperforming DMCE-Global at high coverage in Fig.~\ref{fig:multi_map}. This suggests that selectively extending peer-plan availability can be particularly useful once robots disperse and direct communication becomes sparse. The local map knowledge results in Fig.~\ref{fig:multi_map} further show that exploration time and distributed map knowledge need not improve together. Importantly, PPI+MH can underperform compared to either component individually, particularly in constrained environments. We examine this non-additive interaction in Sec.~\ref{sec:ppi_mh_interaction}.


\subsection{Interaction Between PPI and Multi-hop Propagation}
\label{sec:ppi_mh_interaction}

PPI and MH can each improve exploration, but their benefits are not always additive. At $N=12$ in Tunnel, PPI and MH reduce $T_{95}$ from $106.4$\,s to $99.4$\,s and $89.6$\,s, respectively, whereas PPI+MH increases it to $108.7$\,s. Similar non-additive behavior is observed across the evaluated environments.

To diagnose this interaction, we tested successful MCTS iterations to measure fresh peer plans and PPI exposure along simulated trajectories. In Tunnel, MH increases the mean number of fresh peer plans from $2.30$ to $3.92$, mean PPI exposure from $0.320$ to $0.595$, and the fraction of samples with non-zero exposure from $34.8\%$ to $51.8\%$. Neither reducing the PPI penalty nor restricting PPI construction to directly received plans recovered the combined configuration. This indicates that the degradation is not explained solely by penalty magnitude or by applying PPI to relayed plans.

\begin{figure}[t]
    \centering
    \includegraphics[width=\columnwidth]{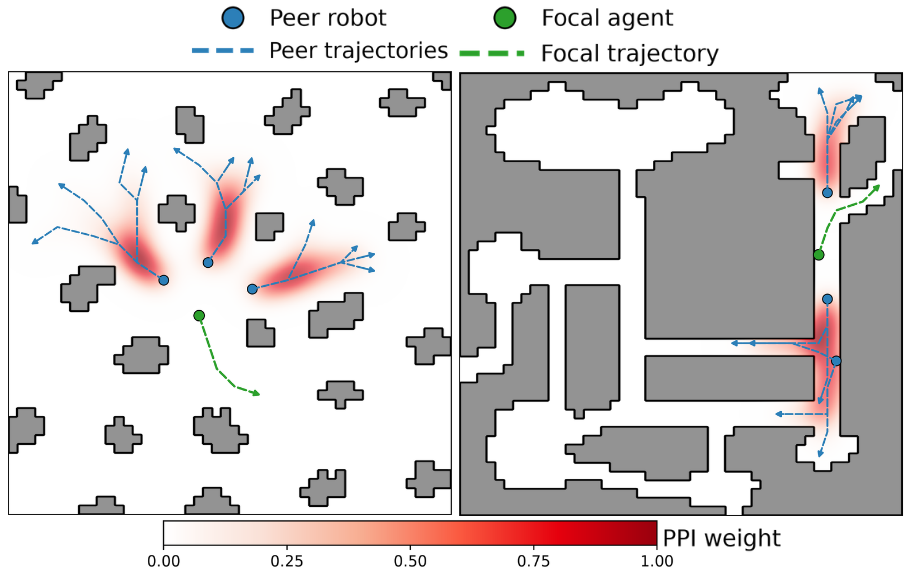}
    \caption{\textbf{Conceptual illustration of environment-dependent peer intent.} The same spatial peer-intent representation can have different implications in open and topologically constrained environments.}
    \label{fig:ppi_topology}
\end{figure}

These results point towards a limitation of purely spatial intent reasoning. As illustrated in Fig.~\ref{fig:ppi_topology}, Euclidean proximity alone cannot distinguish redundant exploration from necessary shared transit through bottleneck areas. When MH expands peer-plan availability, this ambiguity can increase PPI exposure in shared corridors even when robots pursue different regions. Thus, PPI determines how peer information influences local planning, while MH determines how far that information propagates. Their interaction is topology dependent, which is a potential direction for future work. 

\shivaram{Compress this subsection to about 80\% of current.} \saurbh{done}

\subsection{Analysis of Communication and Distributed Information}
\label{sec:multihop_effect}

Multi-hop propagation extends peer-plan availability beyond the immediate communication neighborhood at the cost of additional trajectory exchange. Table~\ref{tab:communication_load} reports the serialized ROS payload rate for $N=5$. PPI+MH requires $230.66$-$342.37$\,kB/s compared with $46.61$-$78.11$\,kB/s for DMCE. Importantly, $26.15$-$54.15\%$ of the source plans processed by PPI+MH are received through relays, confirming that a substantial fraction of planning information originates beyond direct neighbors. 

\begin{table}[t]
\centering
\caption{Trajectory communication for $N=5$ robots. Payload denotes mean serialized payload rate and Relay denotes the fraction of source plans received through relays.}
\label{tab:communication_load}
\footnotesize
\begin{tabular}{lrrr}
\hline
\textbf{Environment} &
\textbf{DMCE} &
\textbf{PPI+MH} &
\textbf{Relay [\%]}$\uparrow$ \\
& \multicolumn{2}{c}{\textbf{Payload [kB/s]}$\downarrow$} & \\
\hline
Open   & 57.74 & 342.37 & 26.15 \\
Forest & 46.77 & 230.66 & 40.18 \\
Urban  & 46.61 & 242.56 & 54.15 \\
Tunnel & 78.11 & 262.18 & 37.96 \\
\hline
\end{tabular}
\end{table}

We also examine how exploration progress is reflected in the local occupancy maps of individual robots. Since Multi-hop propagation exchanges trajectories rather than occupancy grids, greater plan availability does not necessarily imply uniform local map knowledge. At the T95 milestone, Table~\ref{tab:agent_knowledge} reports mean team knowledge, least-informed robot knowledge, and the best-to-worst knowledge gap. In Open and Urban, PPI+MH increases the least-informed robot knowledge by $9.99$ and $11.56$ percentage points while reducing the corresponding gaps by $10.77$ and $11.05$ points. Tunnel shows smaller gains, while Forest exhibits lower and less uniform local knowledge despite faster coverage. These results suggest that propagated planning information and occupancy-map information have different effects. 

\begin{table}[t]
\centering
\caption{Local map knowledge at T95 for $N=5$. Higher Mean K. and Worst K., and lower Gap, are preferable.}
\label{tab:agent_knowledge}
\footnotesize
\begin{tabular}{llccc}
\hline
\textbf{Env.} & \textbf{Method} &
\textbf{Mean K.}$\uparrow$ & \textbf{Worst K.}$\uparrow$ & \textbf{Gap}$\downarrow$ \\
& & \textbf{[\%]} & \textbf{[\%]} & \textbf{[\%]} \\
\hline
\multirow{2}{*}{Open}
& DMCE  & 78.06 & 61.98 & 24.84 \\
& PPI+MH & \textbf{81.32} & \textbf{71.97} & \textbf{14.07} \\
\hline
\multirow{2}{*}{Forest}
& DMCE  & \textbf{72.65} & \textbf{64.10} & \textbf{13.65} \\
& PPI+MH & 69.33 & 56.15 & 21.25 \\
\hline
\multirow{2}{*}{Urban}
& DMCE  & 90.77 & 76.30 & 18.66 \\
& PPI+MH & \textbf{93.26} & \textbf{87.86} & \textbf{7.61} \\
\hline
\multirow{2}{*}{Tunnel}
& DMCE  & 79.18 & 67.24 & \textbf{18.77} \\
& PPI+MH & \textbf{82.43} & \textbf{69.71} & 21.37 \\
\hline
\end{tabular}
\end{table}

\subsection{Ablation on Parameter Sensitivity}
\label{sec:parameter_sensitivity}

We evaluate the sensitivity of PPI to penalty weight $\lambda_{\mathrm{PPI}}$ and spatial spread $\sigma$ in forest environment with $N=5$. Within each experiment set, all methods use the same 10 random seeds, ensuring seed-matched comparisons. Separate experiment sets, including the main evaluation and parameter sweeps, use independently generated seed sets. Tables~\ref{tab:lambda_sweep} and~\ref{tab:sigma_sweep} report the coverage time and overlap.

\begin{table}[t]
\centering
\caption{Effect of $\lambda_{\mathrm{PPI}}$ with $\sigma=1$ in Forest for $N=5$. Lower is better.}
\label{tab:lambda_sweep}
\footnotesize
\begin{tabular}{c | c c | c}
\hline
\textbf{$\lambda_{\mathrm{PPI}}$} & \textbf{T90 [s]}$\downarrow$ & \textbf{T95 [s]}$\downarrow$ & \textbf{Overlap [\%]}$\downarrow$ \\
\hline
0.1 & $73.8 \pm 10.5$ & $104.9 \pm 33.3$ & $46.67 \pm 9.19$ \\
0.3 & $77.3 \pm 7.8$  & $112.9 \pm 21.8$ & $44.60 \pm 8.99$ \\
0.5 & $86.3 \pm 9.8$  & $126.5 \pm 26.6$ & $45.84 \pm 9.03$ \\
1.0 & $\mathbf{77.2 \pm 9.7}$  & $103.9 \pm 15.4$ & $48.51 \pm 7.13$ \\
2.0 & $80.2 \pm 14.2$ & $\mathbf{97.4 \pm 15.3}$ & $\mathbf{43.69 \pm 12.45}$ \\
\hline
\end{tabular}
\end{table}

The effect of $\lambda_{\mathrm{PPI}}$ is non-monotonic. Increasing the peer-intent penalty does not consistently improve either coverage or overlap. $\lambda_{\mathrm{PPI}}=2$ achieves the lowest mean T95 and overlap, whereas intermediate values can perform worse. This indicates that the influence of PPI depends on its interaction with the underlying exploration objective rather than a simple avoidance behavior. We retain $\lambda_{\mathrm{PPI}}=1$ as a common setting throughout the main experiments rather than selecting a separate value for each environment.

\begin{table}[t]
\centering
\caption{Effect of $\sigma$ with $\lambda_{\mathrm{PPI}}=1$ in Forest for $N=5$. Lower is better.}
\label{tab:sigma_sweep}
\footnotesize
\begin{tabular}{c | c c | c}
\hline
\textbf{$\sigma$} & \textbf{T90 [s]}$\downarrow$ & \textbf{T95 [s]}$\downarrow$ & \textbf{Overlap [\%]}$\downarrow$ \\
\hline
0.5 & $76.3 \pm 14.9$ & $\mathbf{96.9 \pm 18.4}$ & $46.96 \pm 7.98$ \\
0.8 & $81.3 \pm 14.7$ & $104.1 \pm 25.8$ & $\mathbf{42.97 \pm 5.95}$ \\
1.0 & $78.1 \pm 10.0$ & $97.2 \pm 17.3$ & $44.65 \pm 11.15$ \\
1.5 & $81.0 \pm 18.2$ & $110.1 \pm 25.8$ & $44.47 \pm 7.19$ \\
2.0 & $75.9 \pm 9.3$  & $101.0 \pm 11.5$ & $47.25 \pm 6.78$ \\
3.0 & $\mathbf{71.5 \pm 8.8}$ & $107.6 \pm 31.4$ & $46.83 \pm 6.23$ \\
\hline
\end{tabular}
\end{table}

Performance is similarly stable over a broad range of spatial scales, with no single $\sigma$ simultaneously minimizing T90, T95, and overlap. The best mean T95 occurs at $\sigma=0.5$, the lowest overlap at $\sigma=0.8$, and the best T90 at $\sigma=3.0$. We therefore use the intermediate setting $\sigma=1$ to avoid environment specific parameters.

\subsection{Hardware Implementation}


\begin{figure*}[t]
    \centering    \includegraphics[width=0.8\textwidth]{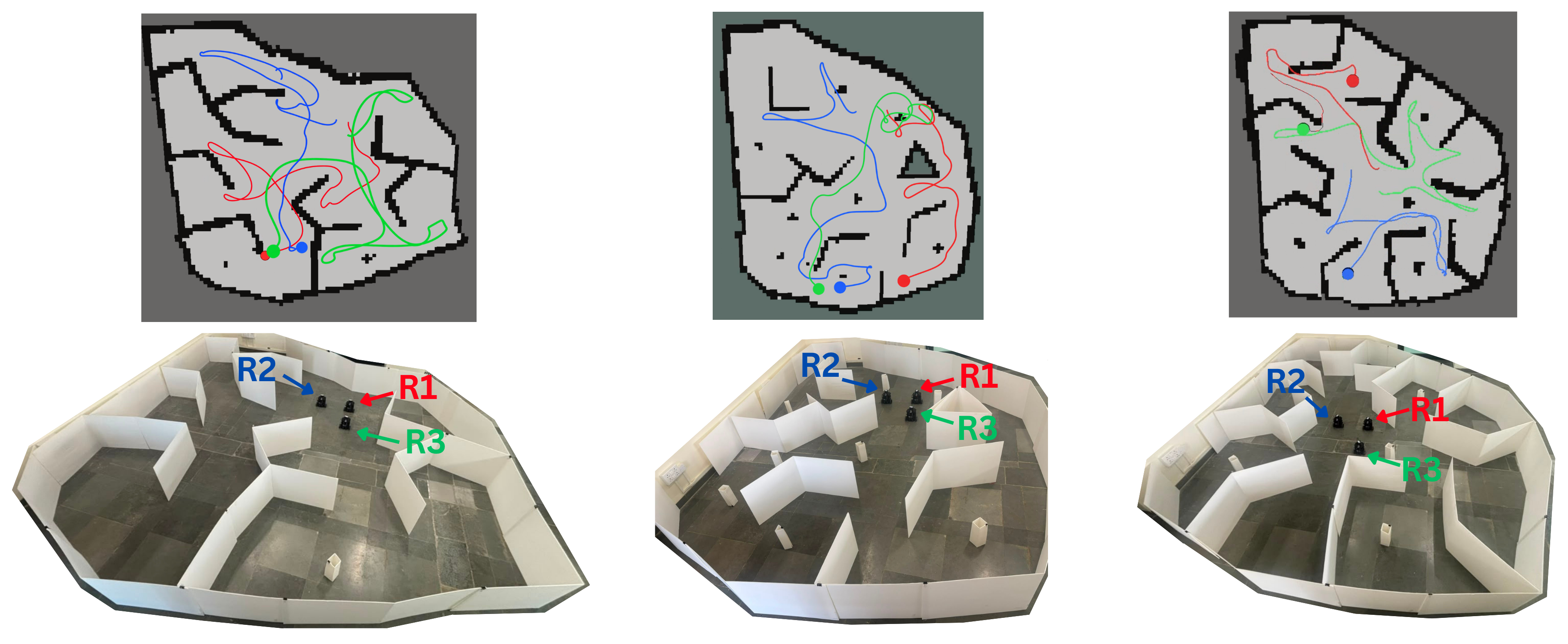}
    \caption{\textbf{Hardware validation.} The proposed framework is deployed on three TurtleBot3 robots in a cluttered indoor environment using onboard LiDAR with an effective sensing range of 1.5\,m and online occupancy mapping. The bottom row shows representative views of the physical testbed with the three robots and their initial positions indicated as R1, R2 and R3, while the top row shows the corresponding occupancy maps and executed trajectories obtained during the exploration.}
    \label{fig:hardware}
\end{figure*}

We validate the proposed framework on a team of three TurtleBot3 robots operating in previously unknown environments. Each robot uses its onboard LiDAR measurements to compute the occupancy maps (restricted to 1.5\,m)
The plans for each robot were generated online using the same decentralized planning framework as in the simulation setup.


The exploration waypoint selected by DMCE+PPI+MH was executed using the standard ROS navigation stack, 
Dynamic Window Approach (DWA) was used for local trajectory execution and obstacle avoidance. Note that the underlying decentralized planner, PPI construction, and Multi-hop propagation remained unchanged from the simulation.

We evaluated the system in multiple indoor environments. Figure~\ref{fig:arena} shows online occupancy-map snapshots from two real-world arenas, together with the trajectories traced by the three robots during decentralized exploration. Figure~\ref{fig:hardware} provides additional hardware-validation scenarios, pairing representative views of the physical testbed with the corresponding occupancy maps and executed trajectories. Across all settings, the robots successfully mapped the environment from onboard sensing, exchanged peer plans online, and executed decentralized exploration, demonstrating transfer of the proposed framework from simulation to physical robots.

\section{Summary and Outlook}

We introduced Probabilistic Peer Intent (PPI) and Multi-hop (MH) peer-plan propagation as complementary extensions to decentralized MCTS for multi-robot exploration. PPI provides a continuous spatial representation of available peer plans, while Multi-hop propagation extends their reach beyond direct communication neighbors. Experiments across environmental structures and team sizes show that both mechanisms can improve exploration individually, while their direct composition is dependent on the environment and scale. Physical experiments with three TurtleBot3 robots demonstrate the feasibility of the framework. The main limitations and corresponding directions for future work are:
\begin{enumerate}
    \item \textbf{Spatial intent representation:} PPI cannot distinguish redundant exploration from exploring shared transit areas that are needed. Reducing the PPI penalty or excluding relayed plans from the PPI field did not resolve this behavior. This motivates the need for topological or goal-level intent.

    \item \textbf{Communication scalability:} Multi-hop propagation increases peer-plan awareness at the cost of additional communication overhead. Hop-aware relay selection, trajectory compression, or communication-budgeted propagation could reduce this overhead while retaining useful non-local information.

    \item \textbf{Physical evaluation:} Our hardware validation is limited to three homogeneous TurtleBot3 robots. Larger robot teams including heterogeneous platforms would be necessary for evaluating scalability and robustness.
\end{enumerate}

\bibliographystyle{IEEEtran}
\bibliography{references}

@inproceedings{bone2023dmce,
  author={S. Bone and L. Bartolomei and F. Kennel-Maushart and M. Chli},
  title={{Decentralised multi-robot exploration using monte carlo tree search}},
  booktitle={2023 IEEE/RSJ International Conference on Intelligent Robots and Systems (IROS)},
  pages={7354--7361},
  year={2023}
}

@inproceedings{cladera2024opportunistic,
  author={F. Cladera and Z. Ravichandran and I. D. Miller and M. A. Hsieh and C. J. Taylor and V. Kumar},
  title={{Enabling large-scale heterogeneous collaboration with opportunistic communications}},
  booktitle={2024 IEEE International Conference on Robotics and Automation (ICRA)},
  pages={2610--2616},
  year={2024}
}

@article{best2019decmcts,
  author={G. Best and O. M. Cliff and T. Patten and R. R. Mettu and R. Fitch},
  title={{Dec-MCTS: Decentralized planning for multi-robot active perception}},
  journal={The International Journal of Robotics Research},
  volume={38},
  number={2-3},
  pages={316--337},
  year={2019}
}

@article{bramblett2023epistemic,
  author={L. Bramblett and N. Bezzo},
  title={{Epistemic planning for multi-robot systems in communication-restricted environments}},
  journal={Frontiers in Robotics and AI},
  volume={10},
  pages={1149439},
  year={2023}
}

@article{bayer2026lowbandwidth,
  author={J. Bayer and J. Faigl},
  title={{Decentralized multi-robot exploration under low-bandwidth communications}},
  journal={Autonomous Robots},
  volume={50},
  number={1},
  pages={7},
  year={2026}
}

@article{jadhav2024wiserx,
  author={N. Jadhav and M. Behari and R. J. Wood and S. Gil},
  title={{WiSER-X: Wireless signals-based efficient decentralized multi-robot exploration without explicit information exchange}},
  journal={arXiv preprint arXiv:2412.19876},
  year={2024}
}

@inproceedings{chiun2025marvel,
  author={J. Chiun and S. Zhang and Y. Wang and Y. Cao and G. Sartoretti},
  title={{MARVEL: Multi-Agent Reinforcement Learning for constrained field-of-View multi-robot Exploration in Large-Scale Environments}},
  booktitle={2025 IEEE International Conference on Robotics and Automation (ICRA)},
  pages={11392--11398},
  year={2025}
}

@inproceedings{meng2025gnnvae,
  author={Y. Meng and N. Majcherczyk and W. Liu and S. Kiesel and C. Fan and F. Pecora},
  title={{Reliable and efficient multi-agent coordination via graph neural network variational autoencoders}},
  booktitle={2025 IEEE International Conference on Robotics and Automation (ICRA)},
  pages={12965--12971},
  year={2025}
}

@inproceedings{yamauchi1997frontier,
  author={B. Yamauchi},
  title={{A frontier-based approach for autonomous exploration}},
  booktitle={Proceedings 1997 IEEE International Symposium on Computational Intelligence in Robotics and Automation (CIRA)},
  pages={146--151},
  year={1997}
}

@article{wang2025survey,
  author={C. Wang and C. Yu and X. Xu and Y. Gao and X. Yang and W. Tang and S. Yu},
  title={{Multi-robot system for cooperative exploration in unknown environments: A survey}},
  journal={arXiv preprint arXiv:2503.07278},
  year={2025}
}

@inproceedings{umari2017rrt,
  author={H. Umari and S. Mukhopadhyay},
  title={{Autonomous robotic exploration based on multiple rapidly-exploring randomized trees}},
  booktitle={2017 IEEE/RSJ International Conference on Intelligent Robots and Systems (IROS)},
  pages={1396--1402},
  year={2017}
}

@inproceedings{yu2021smmr,
  author={J. Yu and J. Tong and Y. Xu and Z. Xu and H. Dong and T. Yang and Y. Wang},
  title={{SMMR-explore: Submap-based multi-robot exploration system with multi-robot multi-target potential field exploration method}},
  booktitle={2021 IEEE International Conference on Robotics and Automation (ICRA)},
  pages={8779--8785},
  year={2021}
}

@article{bartolomei2023fast,
  title={{Fast Multi-UAV Decentralized Exploration of Forests}},
  author={Bartolomei, Luca and Teixeira, Lucas and Chli, Margarita},
  journal={IEEE Robotics and Automation Letters},
  volume={8},
  number={9},
  pages={5576--5583},
  year={2023},
  publisher={IEEE},
  doi={10.1109/LRA.2023.3296037}
}

@article{azpurua2023survey,
  title={{A Survey on the Autonomous Exploration of Confined Subterranean Spaces: Perspectives from Real-World and Industrial Robotic Deployments}},
  author={Azp{\'u}rua, H{\'e}ctor and Saboia, Ma{\'i}ra and Freitas, Gustavo M. and Clark, Lillian and Agha-Mohammadi, Ali-Akbar and Pessin, Gustavo and Campos, Mario F. M. and Macharet, Douglas G.},
  journal={Robotics and Autonomous Systems},
  volume={160},
  pages={104304},
  year={2023}
}

@article{queralta2020collaborative,
  title={{Collaborative Multi-Robot Search and Rescue: Planning, Coordination, Perception, and Active Vision}},
  author={Queralta, Jorge Pena and Taipalmaa, Jussi and Pullinen, Bilge Can and Sarker, Victor Kathan and Gia, Tuan Nguyen and Tenhunen, Hannu and Gabbouj, Moncef and Raitoharju, Jenni and Westerlund, Tomi},
  journal={IEEE Access},
  volume={8},
  pages={191617--191643},
  year={2020}
}

@article{swinton2026evolution,
  title={{The Evolution of Autonomous Systems for Planetary Cave Exploration: A Review}},
  author={Swinton, Sarah and Mitchell, Daniel and Blanche, Jamie and McGookin, Euan and Flynn, David},
  journal={Journal of Field Robotics},
  year={2026}
}

\end{document}